# Raman spectral analysis of mixtures with one-dimensional convolutional neural network

M. Hamed Mozaffari[*] and Li-Lin Tay
National Research Council Canada, Metrology Research Centre, Ottawa, Ontario, Canada
*Corresponding author: email: mhamed.mozaffarimaaref@nrc.gc.ca

**Abstract** - Recently, the combination of robust one-dimensional convolutional neural networks (1-D CNNs) and Raman spectroscopy has shown great promise in rapid identification of unknown substances with good accuracy. Using this technique, researchers can recognize a pure compound and distinguish it from unknown substances in a mixture. The novelty of this approach is that the trained neural network operates automatically without any pre- or post-processing of data. Some studies have attempted to extend this technique to the classification of pure compounds in an unknown mixture. However, the application of 1-D CNNs has typically been restricted to binary classifications of pure compounds. Here we will highlight a new approach in spectral recognition and quantification of chemical components in a multicomponent mixture. Two 1-D CNN models, RaMixNet I and II, have been developed for this purpose. The former is for rapid classification of components in a mixture while the latter is for quantitative determination of those constituents. In the proposed method, there is no limit to the number of compounds in a mixture. A data augmentation method is also introduced by adding random baselines to the Raman spectra. The experimental results revealed that the classification accuracy of RaMixNet I and II is 100% for analysis of unknown test mixtures; at the same time, the RaMixNet II model may achieve a regression accuracy of 88% for the quantification of each component.

## 1 Introduction

In the analytical sciences, Raman spectroscopy has been applied extensively for the identification of unknown compounds. The energy difference between the incident and scattered photons provides a unique fingerprint for identifying different molecules. Raman spectroscopy is fast and non-invasive, and handheld Raman analyzers allow sample analysis to be performed in the field [1]. Raman spectroscopy has been utilized in various industrial processes and areas of science, including planetary exploration of extra-terrestrial targets [2], identification of unknown substances in security applications [3], environmental [4], food sciences [5, 6], classification of bacteria [7], cells [8], tissues [9] and biological materials [10-13]. Recent advances in microfabrication and faster computational resources have transformed Raman spectroscopy from a laboratory-based method to one that is being increasingly deployed in the field [14-16]. Nowadays, first responders use portable Raman analyzers for identifying narcotics, toxins, explosives and precursor chemicals [17, 18]. In contrast, fully automatic and accurate recognition of unknown Raman spectra remains a challenging task.

One approach to solving spectrum identification task is to compare the acquired Raman spectrum with a reference dataset collected under similar experimental conditions. In most cases, the spectrum requires pre-processing (e.g., intensity normalization, noise filtration, and baseline correction) prior to the comparison [19]. Conventional spectral matching approaches are iterative techniques. These techniques identify the most similarities between unknown and reference spectra using criteria such as Euclidean distances, correlation coefficient, and least squares [20]. These statistical methods perform well for pure compounds, and such methods are employed in various commercial software. However, field measurements often encounter complex chemical mixtures with several constituents and often with overlapping bands in Raman spectrum. The analysis of mixtures is much more challenging, especially when the number of distinctive constituents is unknown.

Machine learning (ML) methods have become popular techniques in solving pattern recognition problems. Powerful statistical ML tools such as principal component analysis (PCA) [21], independent component analysis (ICA) [22] and partial least squares regression (PLSR) [23] have been the methods of choice for researchers engaged in more complicated problems in spectral data analysis. Specifically, multi-layer artificial neural networks (ANNs) aim to extract and learn features from large-scale raw datasets. Convolutional neural networks (CNNs) are modern ANNs with deeper architectures consisting of convolutional layers, an essential branch of deep learning (DL). Intricate patterns in high-dimensional data can be quickly discovered using CNNs, reducing the need for manual effort in pre-processing and feature engineering [24].

In recent Raman spectral analysis studies, state-of-the-art DL techniques have become a dominant approach for classifying unknown pure compounds [20, 25-27]. However, only a few studies [1] have investigated the analysis of Raman mixtures. In this study, two one-dimensional (1-D) CNNs (named RaMixNet I and II) are proposed for identifying unknown compounds in the Raman spectra of mixtures. A data augmentation technique is also introduced to create a synthetic dataset of mixtures from a limited number of Raman spectra. RaMixNet I and II are capable of analyzing mixtures with any number of substances. So far, DL usage for mixture analysis has been limited to the qualitative classification of unknown spectra. To the best of our knowledge, this work is the first time that DL models have been used to identify spectral components of a mixture, both qualitatively and quantitatively.

## 2 Materials and Methods

### 2.1 Raman Data Acquisition and Sample Preparation

A complicated case concerning the identification of compounds in a mixture is when the sample consists of many substances with similar Raman spectra. In this experiment, the Raman spectra of four different aromatic compounds, aniline, o-xylene, pyridine and toluene, were acquired for the compounds contained in glass vials. The chemical structures of the four neat compounds are shown in Figure 1A. These chemical structures share a common functional base (the aromatic ring), each with different substituents. The four compounds were selected because of the similarities in chemical structure and their distinctiveness. The compounds are all phenyl-derivatives, thus sharing common vibrational features from the phenyl ring breathing mode. However, all the four selected compounds are distinct with respect to the different substituents which gives each compound unique Raman features that can be used to form the basis-set to build the training data for the testing of our developed deep learning models.

The Raman spectra were measured with a portable Raman analyzer (ReporteR, SciApps, Figure 1B). The ReporteR is equipped with a 70 mW, 785 nm excitation laser; a 2,048-pixel TE-cooled CCD array detector; and a 1800 lines/mm grating. The instrument covers a spectral range of 300-2500 $cm^{-1}$ with a spectral resolution of 12 $cm^{-1}$. Measurements were carried out using an 8 mm glass vial fitted in the liquid sample attachment holder as shown in Figure 1B. This ensured that all measurements were performed with the same sample volume.

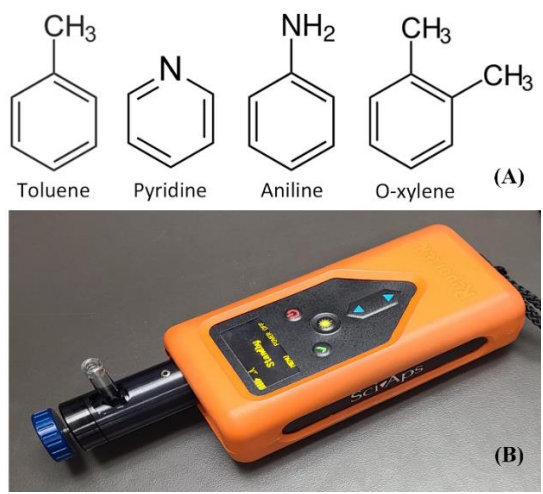

Figure 1 (A) Chemical structures of each pure compound used in this study. (B) Photo of ReporteR, a handheld Raman analyzer. Raman spectra of solutions are acquired with the liquid attachment device as shown.

## 2.2 Training and Test Datasets

Training and validation of a supervised DL model require extensive datasets [24, 28, 29]. The size of each dataset is relative to the complexity of the model and the feature characteristics of the data (e.g., sharp peaks of the Raman spectra) [30]. Six solution mixtures were prepared from the four neat compounds for testing the proposed CNN models. Table 1 gives a summary of the samples, their compositions and the acquisition times.

Table 1 Details of compounds and mixtures used in the test experiments

|    | Sample | Acquisition time |
|----|--------|------------------|
| S1 | 200 μl toluene + 200 μl o-xylene | 2 s |
| S2 | 200 μl pyridine + 200 μl aniline | 0.6 s |
| S3 | 200 μl o-xylene + 200 μl aniline | 0.6 s |
| S4 | 200 μl o-xylene + 200 μl pyridine | 0.6 s |
| S5 | 200 μl o-xylene + 200 μl pyridine + 200 μl Toluene | 1 s |
| S6 | 200 μl toluene + 200 μl aniline | 0.6 s |

Therefore, the training and test datasets contain only four and six Raman spectra, respectively. All possible Raman spectra of the mixtures ranging from binary to ternary to quaternary were prepared by combining the spectra of the four pure compounds using an incremental Raman intensity resolution of 10% to create our training dataset. In general, for $n$ Raman spectra of pure compounds and a Raman intensity resolution of $t$, the total number of possible combinations is $(t^n - 1)$, which in this case, becomes $(10^4 - 1 = 9,999)$ mixtures. There was no pre- or post-processing of data in the subsequent data treatment, including baseline correction or signal denoising and smoothing.

The classification ground truth (GT) labels of the pure compounds were mixed correspondingly in the form of one-hot vectors. The corresponding GT labels are eight-element vectors. The first four elements indicate the class of each pure compound (aniline, o-xylene, pyridine, and toluene, respectively). The second four elements represent the ratio of each pure compound. For instance, a GT label [1, 0, 0, 1, 0.1, 0, 0, 0.7] means that the mixture comprises (10% or 20 μl) aniline and (70% or 140 μl) toluene. Note that there is no limit to the number of compounds in mixtures. Here, four compounds were selected for the sake of simplifying the experiments. Thus, preparing mixtures with a greater number of constituents is possible by tuning the last layer of the DL models.

One approach for increasing the number of data is to augment them by altering each sample in a reasonable range of transformations [31, 32], including shifting, scaling, and rotation. In this study, we also introduced a new data augmentation technique by adding random baselines to the Raman spectra. After generating mixtures by combining pure compounds, data are augmented and normalized using a set of random transformations. Then, we added random baselines, including gaussian, sigmoid, polynomials of degrees from two to five, and exponential functions [33]. Table 2 demonstrates the hyper-parameter values of the baselines employed in this study. Each parameter was selected randomly in a way such that the Raman spectra remained in a reasonable and practical range.

Table 2 Baseline types and parameters utilized in data augmentation

|  | Function Formula | Range of Parameters |
|---|---|---|
| Gaussian Function | $f(x) = Ae^{-(x-\mu)^2/2\sigma^2}$ | Mean ($\mu$) = 1000 to 1500 $cm^{-1}$, Variance ($\sigma^2$) = 800 $cm^{-1}$, Amplitude = 0 to 0.3 |
| Sigmoid Function | $f(x) = A\left(1/(1 + e^{-s \times (x-c)})\right)$ | Slope (s) = 0.001 to 0.03, Centre (c) = 100 to 2400 $cm^{-1}$, Amplitude = 0 to 0.5 |
| Exponential Function | $f(x) = Ae^{-s \times x}$ | Slope (s) = 0.001 to 0.009, Amplitude = 0 to 1 |
| Polynomial Function | $f(x) = \sum_{i=1}^{n} a_i x^i$ | Degree (n) = 1 to 5, $a_i$ = random numbers. $x$ = from 300 to 2500 $cm^{-1}$ mapped from -2.5 to 2.5 by a factor of (5/2201 = 0.00227) |

## 2.3 Multi-label Classification and Regression for Raman Mixtures

One convolutional neural network model takes one input data (e.g., Raman spectrum), extracts the main representative features, and predicts an output based on the extracted features. The output type can be a probability of the input data belonging to a specific class (classification problem) or an estimation value for the input being dependent on the particular application (regression problem). A typical CNN model usually includes several convolutional layers. Convolutional layers aim to find a relationship between features of the previous layer with the input of the next layer to finally create outputs as similar as possible to the ground truth labels provided in training dataset.

Consecutive convolutional layers extract all features in training data from low-level features (e.g., Raman peaks) to high-level features (e.g., rise and fall trends [34]). Activation layers (e.g., Sigmoid or ReLU [20]) between convolutional layers add a non-linearity property to the network for solving non-linear problems [20]. To mitigate computational cost and increase the generalization ability of model, pooling layers are inserted between convolutional layers. Pooling layers between convolutional layers only allow the most significant features to pass from one layer to the next. After several convolutional and pooling layers, which are sometimes called encoder block [35], there may be one or more fully-connected layers. The fully-connected layer usually acts as a classifier that aims to predict the outcome from the features extracted by the encoder block. There are variants of the CNN configurations in the DL literature [20, 24]. However, their core components are similar. For a comprehensive details of convolutional neural networks, refer to [20, 24].

As a taxonomy, classification task can be categorized into several types based on the output shape and type of deep learning models (see Table 3). The identification of different substances in a mixture is a multi-label classification task. For this reason, we specifically designed RaMixNet I for the classification of components in chemical mixtures, as a multi-label convolutional classifier. Figure 2 shows the network architecture of RaMixNet I. We also developed RaMixNet II which can simultaneously solve multi-label classification and regression problems. Figure 3 shows the network architecture of RaMixNet II. The loss functions for the training of RaMixNet models are the binary cross-entropy (BCE). For the case of RaMixNet II, a weighted mean squared error (MSE) loss is used for training the regression path of the model. From Figure 2, the output of RaMixNet I is a binary classifier that determines the presence or absence of compounds in an unknown mixture. Likewise, the output of RaMixNet II has two classifier work simultaneously to predict mixture components and their concentrations. Both RaMixNet I and II have been tested separately by simulated and real mixtures. We can claim that RaMixNet I is a special case of the RaMixNet II when we omit the regression path.

Table 3 A categorization of classification tasks based on the output shape and type.

| Classification Type | Application | Output Type | Ground Truth Labels | Validation Metrics | Example | Note |
|---|---|---|---|---|---|---|
| One-class Classifier | Anomaly detection | One probability value | [1], [1] | Cross-Entropy or MSE | Detecting of a poisoning substance | - |
| Binary Classifier | Distinguishing between two classes of data | One zero or one | [1], [0] | Binary Cross-Entropy or MSE | Distinguishing poisoning from clean substances | - |
| Multi-Class Classifier | Distinguishing between more than two classes of data | One probability value or one class numbers | [1, 0, 0, 0], [0, 1, 0, 0] Or [6], [7] | Sparse Cross-Entropy or MSE | Identifying a pure compound from several | Output can be in the form of a one-hot encoding vector or sparse categorical numbers |
| Multi-Label Classifier | Recognizing several classes of data from one complex data | Vector of binary values | [1, 1, 0, 0], [0, 0, 1, 1] | Binary Cross-Entropy or MSE | Raman Mixture Analysis | Outputs multiple probabilities for each sample |
| Multi-output Multiclass Classifier | Recognizing several classes of data from several complex data | Vector of float values | [8, 4, 1, 0], [1, 6, 3, 4] | Sparse Cross-Entropy or MSE or metrics | Segmentation of a Hyperspectral Raman image | Outputs multiple classes for each sample |

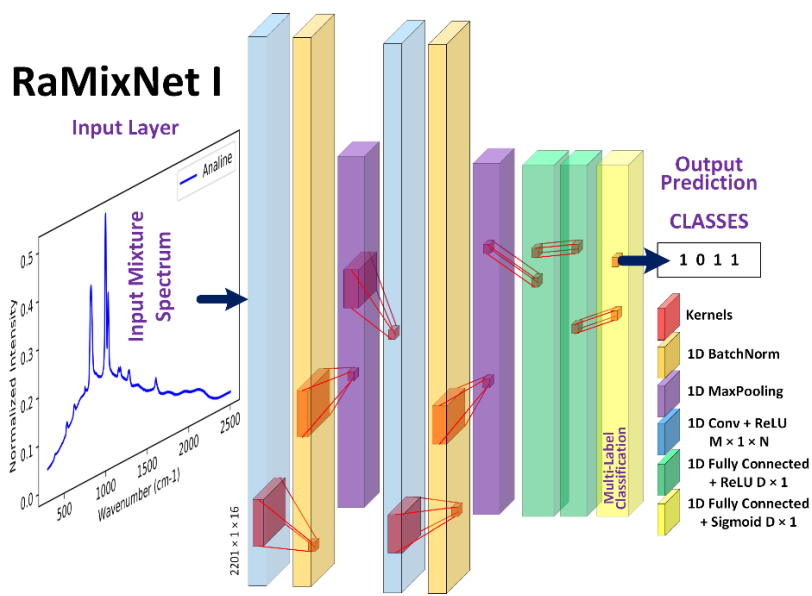

Figure 2 RaMixNet I architecture. M: # of Conv filters, N: Conv filter size, D: # of Neurons in dense layers, C: # of classes. For values of M, N, D, and C refer to hyper-parameter tuning results.

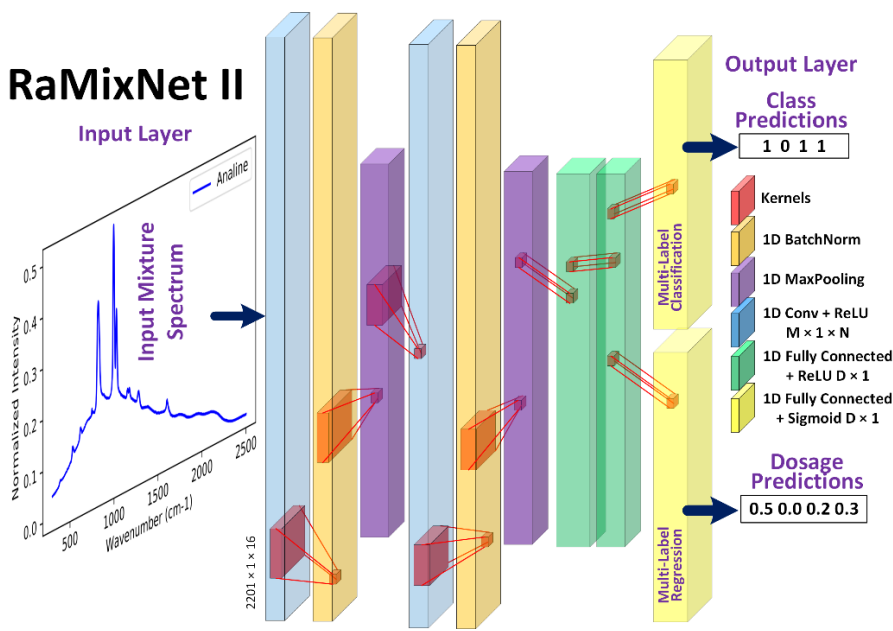

Figure 3 RaMixNet II architecture. M: # of Conv filters, N: Conv filter size, D: # of Neurons in dense layers, C: # of classes. For values of M, N, D, and C refer to hyper-parameter tuning results. RaMixNet II has two outputs with separate activation and loss functions, one for classification and one for regression.

## 3 Experimental Results

Both RaMixNet I and II were implemented using the publicly available TensorFlow library on a 64-bit Windows PC with 20 CPU cores, 192GB of memory, and an NVidia TITAN RTX GPU. An analytical assessment of our classification results using criteria derived from confusion matrices was employed. The task here is a multi-label classification and not binary classification, therefore, there are four confusion matrices, one per mixture component. For the case of multi-label regression analysis, values for the Mean Square Error loss and the regression accuracy are reported. Note that the outputs of the CNN models are probability values, and for this reason, the regression results are reported as approximations. Besides, the coefficient of determination ($R^2$ score) was used to measure how well the regression line approximates the actual data.

We report the evaluation results of the RaMixNet I and II in Table 4 and Table 5, respectively. From Table 4, RaMixNet I could classify all components of the six mixtures with a classification accuracy of 100%. Conducting the same experiment for RaMixNet II indicates the powerful capability of this model with regard to both multi-label classification and regression tasks. From the evaluation results of Table 5, although the classification accuracy of RaMixNet II for mixture S6 was 100%, the regression path of the model detected 58 µl for Pyridine, whereas the solution S6 has no Pyridine component. The reason might come from the fact that the two sharp characteristic peaks of the Pyridine are relatively stronger than other peaks. While for other components, there are other significant peaks for distinguishing. For this reason, it is natural that RaMixNet II detects Pyridine with a small concentration for mixture S6.

If the goal of an investigation is only the identification of substances in an unknown mixture, using RaMixNet I, which has fewer parameters and a faster inference speed is more suitable. This choice is even more appropriate for the case of using handheld Raman spectrometers. By contrast, RaMixNet II is the preferred model for cases where both qualitative and quantitative analysis of mixtures are the desired goal.

Table 4 Evaluation results of RaMixNet I. Inference values are the output of the sigmoid function in the last layer of RaMixNet I

|    | True Components              | Predicted Components         | Classification Analysis |    |    |    |    |      |      |      |      |          |      |
|----|------------------------------|------------------------------|-------------------------|----|----|----|----|------|------|------|------|----------|------|
|    |                              |                              |                         | TP | FN | FP | TN | TPR  | TNR  | PPV  | NPV  | Accuracy | F1   |
| S1 | O-Xylene, Toluene            | O-Xylene, Toluene            | Aniline                 | 3  | 0  | 0  | 3  | 1.00 | 0.50 | 1.00 | 1.00 | 100%     | 1.00 |
| S2 | Aniline, Pyridine            | Aniline, Pyridine            | O-Xylene                | 4  | 0  | 0  | 2  | 1.00 | 0.33 | 1.00 | 1.00 | 100%     | 1.00 |
| S3 | Aniline, O-Xylene            | Aniline, O-Xylene            | Pyridine                | 3  | 0  | 0  | 3  | 1.00 | 0.50 | 1.00 | 1.00 | 100%     | 1.00 |
| S4 | O-Xylene, Pyridine           | O-Xylene, Pyridine           | Toluene                 | 3  | 0  | 0  | 3  | 1.00 | 0.50 | 1.00 | 1.00 | 100%     | 1.00 |
| S5 | O-Xylene, Pyridine, Toluene  | O-Xylene, Pyridine, Toluene  |                         |    |    |    |    |      |      |      |      |          |      |
| S6 | Aniline, Toluene             | Aniline, Toluene             |                         |    |    |    |    |      |      |      |      |          |      |

Table 5 Evaluation results of RaMixNet II. Inference values are the output of the two sigmoid functions in the last layer of RaMixNet II. The four GT vector components are Aniline, O-Xylene, Pyridine, and Toluene, respectively, for both classification and regression vectors.

|    | Components                   | Ground truth labels |              | RaMixNet II Predictions |                          | Detection |          |          |         |
|----|------------------------------|---------------------|--------------|-------------------------|--------------------------|-----------|----------|----------|---------|
|    |                              | Classes             | Dosages      | Classes                 | Dosages                  | Aniline   | O-xylene | Pyridine | Toluene |
| S1 | O-Xylene, Toluene            | [0, 1, 0, 1]        | [0, 1, 0, 1] | [0, 1, 0, 1]            | [0.01, 0.97, 0.08, 0.96] | 2 μl      | 194 μl   | 16 μl    | 192 μl  |
| S2 | Aniline, Pyridine            | [1, 0, 1, 0]        | [1, 0, 1, 0] | [1, 0, 1, 0]            | [0.88, 0.15, 0.94, 0.16] | 176 μl    | 30 μl    | 188 μl   | 32 μl   |
| S3 | Aniline, O-Xylene            | [1, 1, 0, 0]        | [1, 1, 0, 0] | [1, 1, 0, 0]            | [0.90, 0.90, 0.65, 0.17] | 180 μl    | 180 μl   | 130 μl   | 34 μl   |
| S4 | O-Xylene, Pyridine           | [0, 1, 1, 0]        | [0, 1, 1, 0] | [0, 1, 1, 0]            | [0.03, 0.55, 0.64, 0.13] | 6 μl      | 110 μl   | 128 μl   | 26 μl   |
| S5 | O-Xylene, Pyridine, Toluene  | [0, 1, 1, 1]        | [0, 1, 1, 1] | [0, 1, 1, 1]            | [0.04, 0.57, 0.68, 0.66] | 8 μl      | 114 μl   | 136 μl   | 132 μl  |
| S6 | Aniline, Toluene             | [1, 0, 0, 1]        | [1, 0, 0, 1] | [1, 0, 0, 1]            | [0.85, 0.07, **0.29**, 0.85] | 170 μl | 14 μl | **58 μl** | 170 μl |

|                         | TP | FN | FP | TN | TPR  | TNR  | PPV  | NPV  | Accuracy | F1   |
|-------------------------|----|----|----|----|------|------|------|------|----------|------|
| Classification Analysis |    |    |    |    |      |      |      |      |          |      |
| Aniline                 | 3  | 0  | 0  | 3  | 1.00 | 0.50 | 1.00 | 1.00 | 100%     | 1.00 |
| O-Xylene                | 4  | 0  | 0  | 2  | 1.00 | 0.33 | 1.00 | 1.00 | 100%     | 1.00 |
| Pyridine                | 3  | 0  | 0  | 3  | 1.00 | 0.50 | 1.00 | 1.00 | 100%     | 1.00 |
| Toluene                 | 3  | 0  | 0  | 3  | 1.00 | 0.50 | 1.00 | 1.00 | 100%     | 1.00 |

# 4 Conclusion and Discussion

In this study, two one-dimensional convolutional neural network models, RaMixNet I and II, were developed for the qualitative and quantitative analysis of Raman spectra of unknown mixtures. RaMixNet I proved to be an expert network for identification of substances in an unknown mixture. At the same time, RaMixNet II had an additional feature that enabled the architecture to perform classification and regression simultaneously. In other words, RaMixNet II can identify unknown components in a mixture and provide an estimation of the quantity of each unknown compound. The proposed deep learning models were trained and validated using Raman spectra of pure, binary, ternary, and quaternary combinations of four pure compounds. The Raman spectra of real mixtures were utilized to evaluate the DL architectures. Training data were augmented using standard transformations used in spectroscopy, including vertical shifting and scaling. The Raman spectra of mixtures were also augmented by adding random baselines with different distributions (e.g., Gaussian, exponential, and polynomial).

Our experiments were based on the use of four compounds. However, the proposed method may be extended to analyze an arbitrary number of substances in a mixture. That is, it should be possible to consider any sample mixture as a linear combination of many pure compounds at different compositional ratios. In practice, the number of substances in a mixture should be not more than a dozen. Therefore, if we make all possible linear combinations of pure compounds and their ratios with a specific Raman intensity resolution, RaMixNet models can be fitted to the dataset to identify the components of all possible mixtures for that application, qualitatively and quantitatively. The experimental results demonstrated the accuracy and robustness of RaMixNet models in solving classification and regression tasks for Raman spectra of chemical mixtures. RaMixNet I and II achieved identification accuracies of 100%. At the same time, RaMixNet II could reach a regression accuracy of 88%.

Supervised DL architectures are data-hungry networks whereby feeding them more data translates into higher accuracy. Training one-dimensional deep convolutional neural networks on a large reference dataset with hundreds of simulated mixtures is a new direction for spectral analysis. RaMixNet models are representative of successful deep learning models and are considered to be in their first phases of development. Further investigations regarding the performance for different complex spectral datasets are planned.